\documentclass{article}
\usepackage{graphicx} 
\usepackage{subcaption}
\usepackage{listings}
\usepackage{xcolor}
\usepackage{url}
\usepackage{authblk}
\usepackage{bookmark}
\usepackage{rotating}
\usepackage{lscape}
\usepackage{float}
\PassOptionsToPackage{table}{xcolor}
\usepackage[table]{xcolor}
\usepackage{longtable}
\usepackage[title]{appendix}
\usepackage[font=small,labelfont=bf]{caption}
\usepackage{adjustbox}
\usepackage{booktabs}
\usepackage{changepage}
\usepackage{array}
\usepackage{multirow}
\usepackage{MnSymbol}
\usepackage{placeins}
\usepackage{amsmath}
\usepackage{gensymb}
\usepackage{fontawesome5}
\usepackage{hyperref}
\usepackage{tikz}
\usepackage{datetime}
\usepackage[margin=1in]{geometry}

\usepackage[backend=biber,style=numeric,hyperref=true,backref=true,bibencoding=utf8,doi=true,url=true,maxcitenames=1]{biblatex}

\title{EFPI: Elastic Formation and Position Identification in Football (Soccer) using Template Matching and Linear Assignment}
\author[1,2]{Joris Bekkers}

\affil[1]{U.S. Soccer Federation}
\affil[2]{UnravelSports}

\date{\monthname\ \the\year}

\begin{document}

\maketitle

\begin{abstract}
\noindent
Understanding team formations and player positioning is crucial for tactical analysis in football (soccer). This paper presents a flexible method for formation recognition and player position assignment in football using predefined static formation templates and cost minimization from spatiotemporal tracking data, called EFPI. Our approach employs linear sum assignment to optimally match players to positions within a set of template formations by minimizing the total distance between actual player locations and template positions, subsequently selecting the formation with the lowest assignment cost. 

To improve accuracy, we scale actual player positions to match the dimensions of these formation templates in both width and length. While the method functions effectively on individual frames, it extends naturally to larger game segments such as complete periods, possession sequences or specific intervals (e.g. 10 second intervals, 5 minute intervals etc.). Additionally, we incorporate an optional stability parameter that prevents unnecessary formation changes when assignment costs differ only marginally between time segments.
\newline
\newline
\noindent
EFPI is available as open-source code through the \textbf{\textit{unravelsports}} \cite{unravelsports} Python package.
\end{abstract}

\section{Introduction}
\label{intro}
The analysis of team formations and player positioning has become increasingly important in modern football analytics, driven by the widespread availability of high-resolution spatiotemporal tracking data. Understanding how teams organize tactically and how players fulfill different roles provides valuable insight to coaches, analysts, and researchers.

Traditional approaches to formation analysis often rely on manual annotation, rudimentary static formation labels (i.e. the label is not changed until a substitution is made), or simplified heuristics. The challenge lies in automatically interpreting the continuous dynamic nature of player movements and translating these into meaningful tactical descriptions. Players rarely maintain perfect formation shapes during play, requiring robust methods that can handle positional variation while still identifying underlying tactical structures.

Early computational approaches to this problem recognized the potential of optimization techniques. Wei et al.\cite{wei2013large} used the Hungarian Algorithm for large-scale formation analysis in soccer, demonstrating that assignment algorithms could effectively match players to formation positions. Their work established the foundation for treating formation recognition as an optimal assignment problem, where the goal is to minimize the total cost of assigning players to predefined tactical roles. Poppeliers (2025)\cite{poppeliers25} points out that this approach can produce illogical results since it focuses only on minimizing the cost function.

Subsequent research has explored various aspects of tactical analysis. Shaw \& Glickman (2019) \cite{shaw2019} developed dynamic analysis methods for understanding team strategy in professional football in five minute increments, emphasizing the temporal aspects of tactical behavior. More recently, Kim et al. (2022)\cite{kim2022} introduced change-point detection methods for identifying formation and role transitions during matches, highlighting the importance of capturing tactical evolution over time. Pleuler (2024)\cite{pleuler24} showcased an XGBoost classifier to automatically detect tactical roles for each player on a frame-by-frame basis during a match. Their supervised classification approach uses ground truth labels from starting lineups to train the model under the assumption that most players occupy their assigned roles for a majority of the game. They encode each player's spatial relationship to their teammates by measuring the number of visible teammates across different angular orientations, represented as a waveform that serve as input for the classification.

Our work extends this line of research by presenting a relatively simple, transparent, easily adaptable, yet comprehensive approach to dynamic position labeling that combines the assignment optimization principles established by Wei et al. (2013)\cite{wei2013large} with practical considerations for automated tactical analysis. We employ the linear sum assignment problem \cite{kuhn1955hungarian} \cite{scipy2020} to optimally match players to positions for a set of template formations, by minimizing the total distance between actual player locations and the template positions, and subsequently choosing the lowest cost formation template. Prior to computing the cost we scale the actual player positions to match the formation dimensions to make the method more robust for real-world applications, this (largely) resolves the illogical assignment problem pointed out by Poppeliers (2025)\cite{poppeliers25}. By doing this we are able to instantly identify formation labels while concurrently assigning position labels to individually players. Our approach is not constrained to a single time interval and can be employed on a frame-by-frame basis, per possession, or any custom time interval (e.g. 10 seconds, 5 minutes) or per full period of play.
\newline

\noindent
The full implementation of EFPI is available as open-source code through the \textbf{\textit{unravelsports}} \cite{unravelsports} Python package, introduced by Bekkers \& Sahasrabudhe (2024) \cite{bekkers2024graph}.

\section{Methodology}
\label{app:position-labels}

\subsection{Formation Templates}
We compute the cost of assigning each team a formation $k$ and corresponding position labels $p_{k,j}$ to players using Formula~\ref{eq:linear-sum-assignment} across a set of 65 predefined static formation templates $K$. These templates are derived from the \textit{mplsoccer} Python library (version 1.5.0) \cite{mplsoccer}, with three representative examples shown in Figure \ref{fig:3-forms} and the complete set depicted in Appendix \ref{app:form}.

Our template collection includes formations for 8, 9, and 10 outfield players to accommodate situations with missing players due to red cards or injuries. The set also includes attacking-oriented formations such as 1432 and 2233. We exclude goalkeepers from our analysis, assuming the data provider's position labels for this role are accurate across all frames.

\begin{figure}[H]
    \centering
    \includegraphics[width=0.90\linewidth]{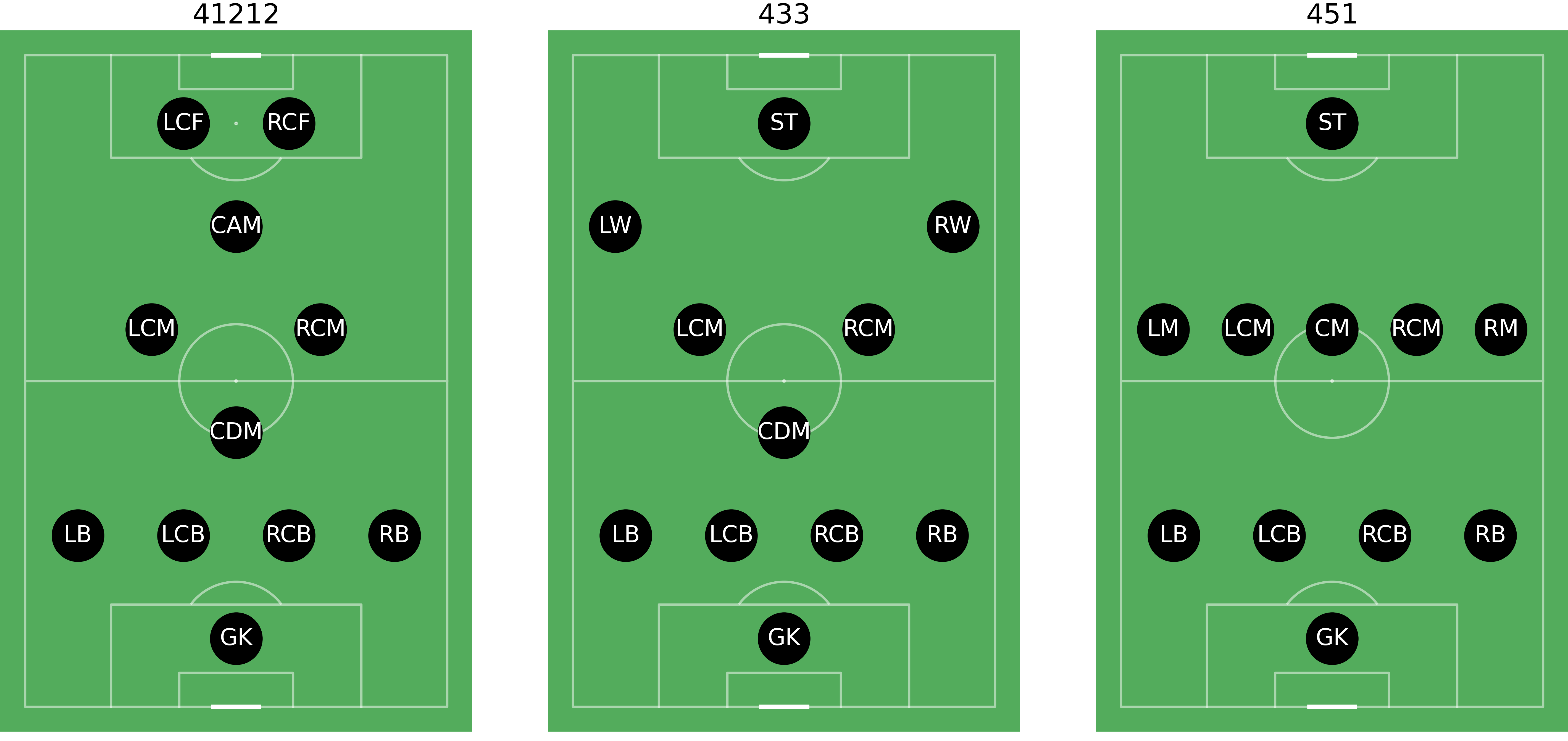}
    \caption{Three example formation templates from \textit{mplsoccer}}
    \label{fig:3-forms}
\end{figure}

\noindent
In Formula~\ref{eq:linear-sum-assignment}, $C_{i,j}$ is the distance between the location ($p_i$) of a player $i$ and the location ($p_j$) of position label $j$. The objective is to assign each player a position label for a total minimum cost (i.e. distance from current position to static formation position). $X$ is a binary matrix where $X_{i,j}$ equals 1 if player $i$ is assigned to position label $j$. We do this calculation for each of the 65 templates, choosing the lowest cost formation template and subsequently assigning each player their position label from this lowest cost template.

\begin{equation}
    \min_{k \in K} \left[ \min \sum_{i} \sum_{j} C_{i,j} X_{i,j} \right]
    \label{eq:linear-sum-assignment}
\end{equation}

\begin{figure}[H]
    \begin{subfigure}[b]{0.47\textwidth}
        \centering
        \includegraphics[width=1\linewidth]{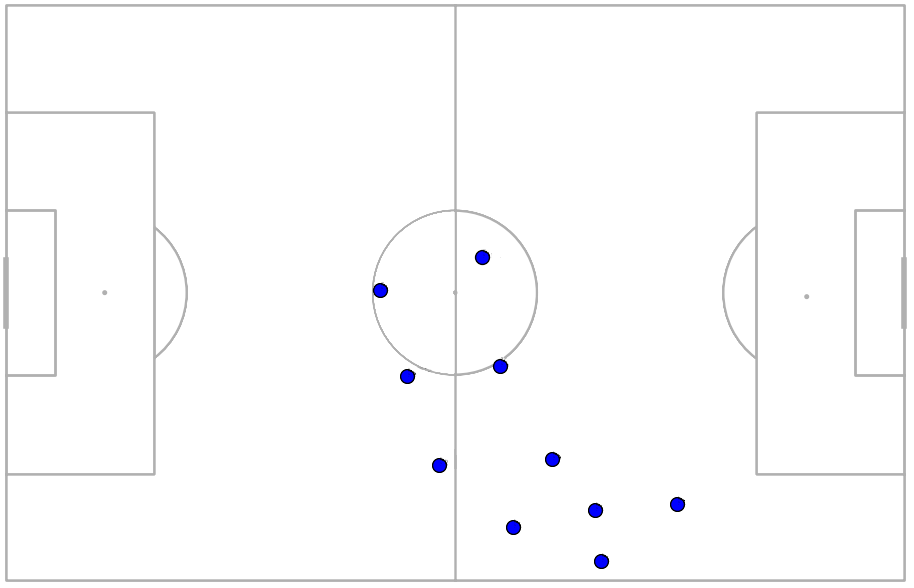}
        \caption{Unlabeled original player positions}
        \label{fig:position11}
    \end{subfigure}
    \hfill
    \begin{subfigure}[b]{0.47\textwidth}
        \centering
        \includegraphics[width=1\linewidth]{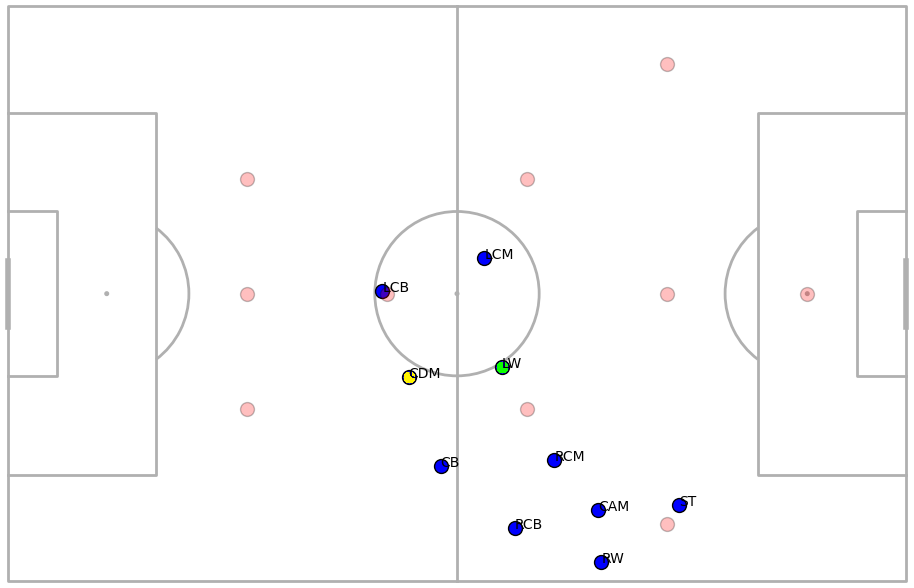}
        \caption{31231 the minimum cost formation before scaling}
        \label{fig:position2}
    \end{subfigure}
    
    \vspace{0.5cm} 
    
    \begin{subfigure}[b]{0.47\textwidth}
        \centering
        \includegraphics[width=1\linewidth]{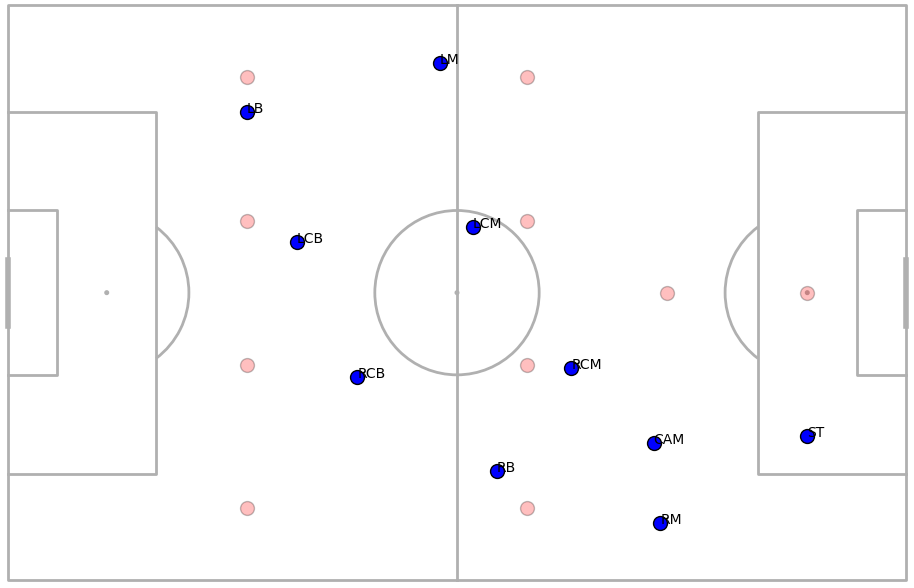}
        \caption{4411 the minimum cost formation after scaling}
        \label{fig:position3}
    \end{subfigure}
    \hfill
    \begin{subfigure}[b]{0.47\textwidth}
        \centering
        \includegraphics[width=1\linewidth]{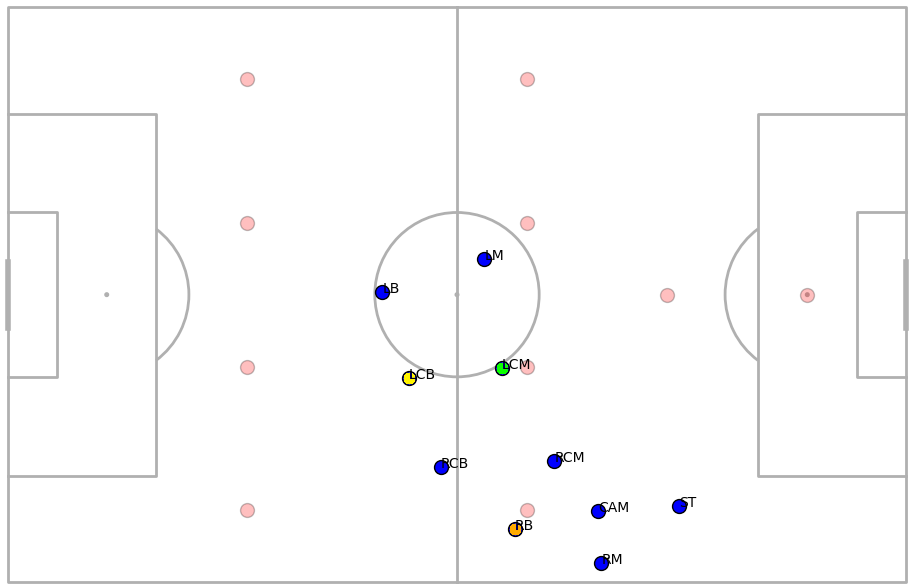}
        \caption{4411 with original player positions}
        \label{fig:position4}
    \end{subfigure}
    
    \caption{Dynamic Position Labeling using the Hungarian algorithm to fit the lowest cost predefined static formation and its position labels to a scaled frame of tracking data. Direction of play is left to right.}
    \label{fig:position-total}
\end{figure}

\noindent
To illustrate our approach we depict a frame of unlabeled original player positions (attacking left to right) in Figure~\ref{fig:position11}. In Figure~\ref{fig:position2} we show the minimum cost formation template for this frame of tracking data, namely "31213". 

It is clear that the associated position labels are incorrect because what should logically be identified as a Centerback (CB) (\tikz[baseline=-0.5ex]\draw[fill=yellow, draw=black, line width=0.3pt] (0,0) circle (2.5pt); in Figure~\ref{fig:position2}) is assigned the Defensive Midfielder (DM) label, and what should logically be identified as a Left (Central) Midfielder (\tikz[baseline=-0.5ex]\draw[fill=green, draw=black, line width=0.3pt] (0,0) circle (2.5pt); in Figure~\ref{fig:position2}) is identified as a Left Winger (LW). 

These assignment issues occur because the actual formation (see Figure~\ref{fig:position11}) is not at the same scale as the formation templates. To resolve this issue, we scale the original player positions to the maximum width and length of all formation templates (see Figure \ref{fig:position3}). We subsequently calculate the assignment cost for each formation template $k$ to these scaled player positions instead. Figure~\ref{fig:position4} shows the minimum-cost formation and the associated original player positions. 

Figure~\ref{fig:position4} shows the minimum cost formation assigned after scaling. It is now identeified as a "4411" with more accurate labels assigned to \tikz[baseline=-0.5ex]\draw[fill=yellow, draw=black, line width=0.3pt] (0,0) circle (2.5pt); (Left Center Back), \tikz[baseline=-0.5ex]\draw[fill=green, draw=black, line width=0.3pt] (0,0) circle (2.5pt); (Left Central Midfielder). Additionally we see that \tikz[baseline=-0.5ex]\draw[fill=orange, draw=black, line width=0.3pt] (0,0) circle (2.5pt); has been more accurately assigned the Right Back (RB) label, compared to the Right Center Back label it obtained in the unscaled variant.

\subsection{Segmentation}
\label{segmentation}
In the above example, we compute the least cost formation template on a single frame of positional tracking data. However, this same approach can be extended to assign formation templates to longer segments of games, such as a full period of play, a possession sequence, or a fixed time window. 

To accomplish this, we replace the single-frame player position $p_i$ with the average location of a player during a segment. Specifically, we compute the average position for player $i$ over a segment ($m$) that contains $q$ frames as:

\begin{equation}
\bar{p}_{i,m} = \frac{1}{N} \sum_{q=1}^{N} p_{i, q}
\end{equation}

where $N$ represents the number of frames in the segment and $p_{i, q}$ is the player's position at frame $q$. This averaged position $\bar{p}$ captures the player's typical location throughout the segment, providing a more stable basis for formation assignment.

To ensure meaningful formation assignments, we partition these segments by attacking and defending periods before applying the template matching algorithm.

\subsection{Stability}
\label{stability}
Given the non-probabilistic nature of our approach, it is relatively stable when assigning formations. However, the availability of 65 formation templates can introduce unwanted noise when analyzing smaller segments or single frames, since the algorithm always selects the cheapest formation and some formations do not differ that much in layout. This becomes particularly problematic during transitional periods, such as when teams move up the pitch and shift between formations. This instability could reduce the method's effectiveness for practical applications depending on the use case. 

To address these fluctuations, we introduce an optional stability factor $\epsilon$. This parameter requires that the cost of assigning a new formation template at time $t$ must be at least $\epsilon$ percent lower than the cost of the template assigned at time $t-1$ before a change in formation is accepted. Formally, a template change is permitted only when:

\begin{equation}
\frac{C_{t-1} - C_t}{C_t} > \epsilon
\end{equation}
where $C_t$ represents the cost of the best-fitting template at time $t$, and $C_{t-1}$ is the cost of the currently assigned template from the previous time step. This stability constraint reduces unnecessary template switching while preserving genuine formation changes.

\begin{figure}[H]
  \centering
  \begin{subfigure}[b]{0.90\linewidth}
      \centering
      \includegraphics[width=\linewidth]{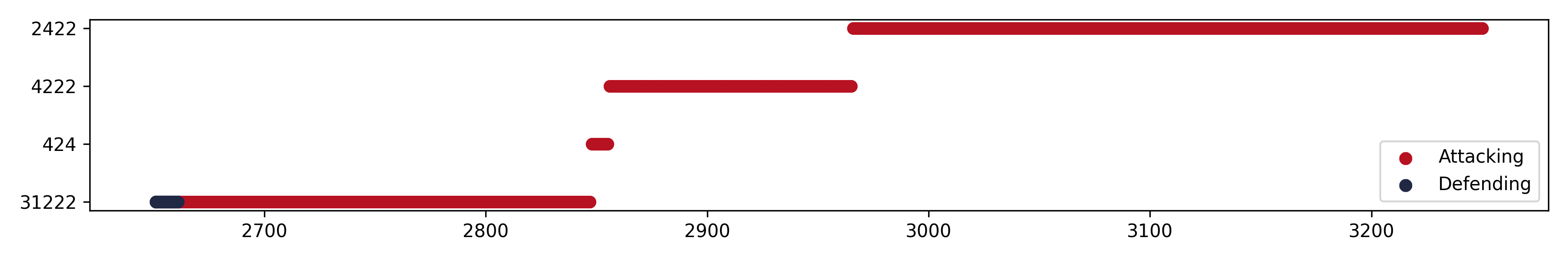}
      \caption{Without a stability parameter}
      \label{fig:position1a}
  \end{subfigure}
  
  \vspace{0.5cm}
  
  \begin{subfigure}[b]{0.90\linewidth}
      \centering
      \includegraphics[width=\linewidth]{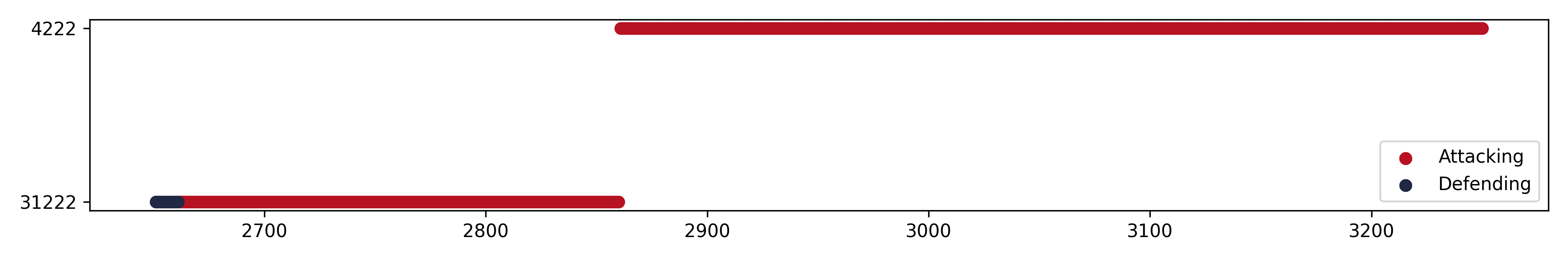}
      \caption{With the stability parameter $\epsilon = 0.1$}
      \label{fig:position1b}
  \end{subfigure}
  
  \caption{A 24 second sequence of play for a single team transitioning from defense to attack, showing the effect of the stability parameter on formation template assignment.}
  \label{fig:fluct}
\end{figure}

\noindent
Figure \ref{fig:fluct} illustrates this phenomenon through a 24-second sequence where a team transitions from defense to attack. As shown in Figure \ref{fig:position1a}, without any stability constraints, the formation assignments changes more frequently throughout the sequence, going from "31222", to "424", to "4222", and "2422". In contrast, Figure \ref{fig:position1b} demonstrates how introducing a stability parameter ($\epsilon$=0.1) produces more stable results. Here, we again start at "31222" and only change to "4222" when the cost of changing is at least 10\% lower. Because the assignment of the "2422" template falls within this 10\% we ultimately stay at "4222".

\subsection*{Example}

\begin{figure}[H]
    \centering
    \includegraphics[width=.9\linewidth]{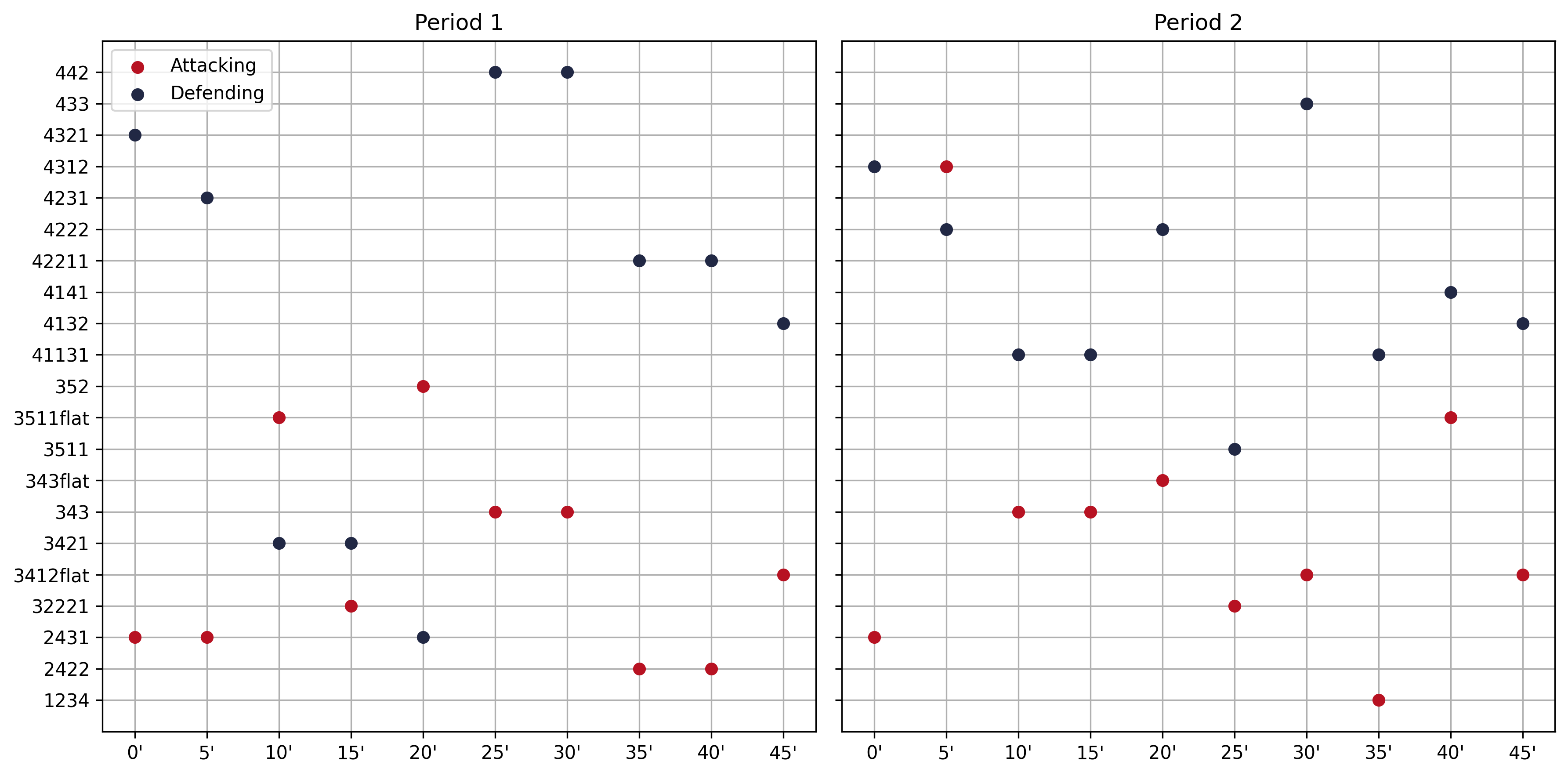}
    \caption{Portugals' formations against Switzerland during World Cup 2022}
    \label{fig:team}
\end{figure}

Depending on the data source Portugal, in their 6-1 victory over Switzerland during the 2022 World Cup, played in a "4312"\footnote{\href{https://www.fotmob.com/matches/switzerland-vs-portugal/1voqy6}{FotMob: Switzerland vs Portugal}} or "433"\footnote{\href{https://www.espn.co.uk/football/match/_/gameId/633842/switzerland-portugal}{ESPN Match Report: Switzerland vs Portugal}} with Bruno Fernandes either playing as a Central Attacking Midfielder (CAM) or as a Right Winger (RW).

EFPI can help us better understand Portugals formation(s) and Bruno Fernandes' position(s), by assigning them to segments, instead of the match as a whole. Figure \ref{fig:team} depicts EFPI applied to Portugal, in 5-minute segments, spread across two halves, splitting attacking and defending situations, $\epsilon$ set to 0.0, and dropping the least occurring players (in the case of substitutions leading to more than 11 unique players in a segment). We can identify a clear separation between attacking and defending formations, where the attacking formations are heavily focused on 3 and 2 at the back, while defending is primarily done in setups with 4 defenders. However, the "433" and "4312" are only assigned once and twice respectively. 

\begin{figure}[H]
    \centering
    \includegraphics[width=.9\linewidth]{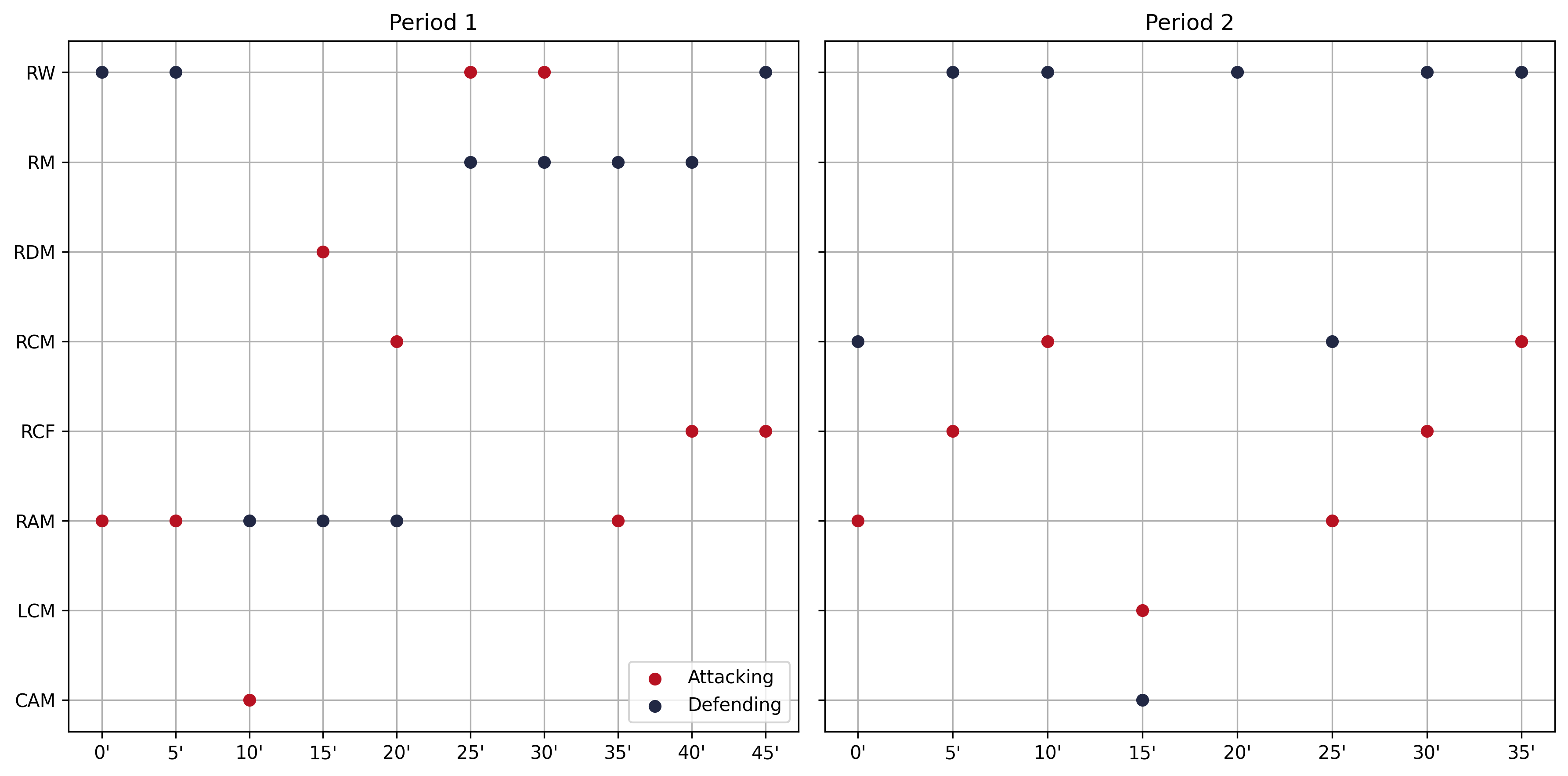}
    \caption{Bruno Fernandes' playing positions against Switzerland.}
    \label{fig:player}
\end{figure}

Figure \ref{fig:player} depicts Bruno Fernandes' assigned position labels during the same intervals. It is immediately clear that in defense he played mostly as a Right Winger (RW), a Right Attacking Midfielder (RAM), or a Right Midfielder (RM). In attack he played similar positions, but sometimes he also showed up as a Right Central Forward (RCF) in a two-man attack.

\subsection{Open Source}
\label{open-source}
The full implementation of the the EFPI algorithm is available as open-source code through the \newline \textbf{\textit{unravelsports}} \cite{unravelsports} Python package. It leverages the \textit{\textbf{Kloppy}}\cite{kloppy} Python package, and as a result works for 9 different tracking data providers, namely HawkEye, Metrica, PFF, SecondSpectrum, Signality, SkillCorner, Sportec, StatsPerform and Tracab. 

Codeblock \ref{lst:efpi-config} depicts a simple Python implementation using Bassek et al. (2025)\cite{bassek2025} openly available positional data.

\begin{center}
\begin{minipage}{0.9\textwidth}
\begin{lstlisting}[caption={EFPI Model Configuration and Training}, captionpos=b, label={lst:efpi-config}]
from kloppy import sportec
from unravel.soccer import KloppyPolarsDataset, EFPI

# load open data using kloppy
kloppy_dataset = sportec.load_open_tracking_data()

# convert to Polars dataframe
kloppy_polars_dataset = KloppyPolarsDataset(
    kloppy_dataset=kloppy_dataset
)

# fit EFPI
model = EFPI(dataset=kloppy_polars_dataset)
model.fit(
    # Default 65 formations, or specify a subset (e.g. ["442", "433"]
    formations=None,  
    # 5-minute intervals, or specify "possession", "period", "frame" etc.
    every="5m",  
    substitutions="drop",
    change_threshold=0.1,
    change_after_possession=True,
)
\end{lstlisting}
\end{minipage}
\end{center}

\newpage
\begin{appendices}
\section{Formation Templates}
\label{app:form}
\begin{figure}[H]
    \centering
    \includegraphics[width=0.86\linewidth]{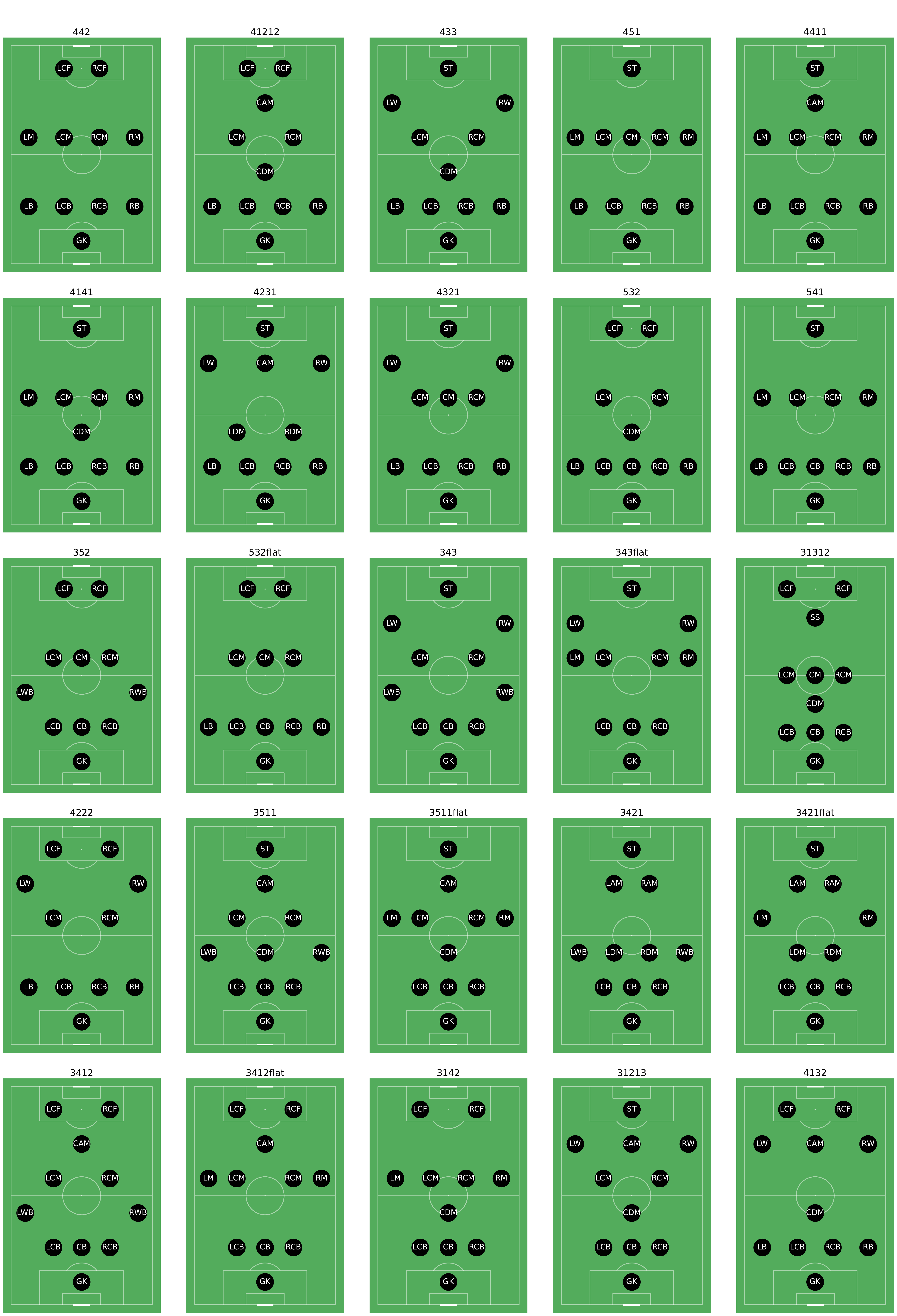}
    \caption{\textit{mplsoccer} Formation Templates 1 - 25}
    \label{fig:positiona}
\end{figure}
\begin{figure}[H]
    \centering
    \includegraphics[width=0.86\linewidth]{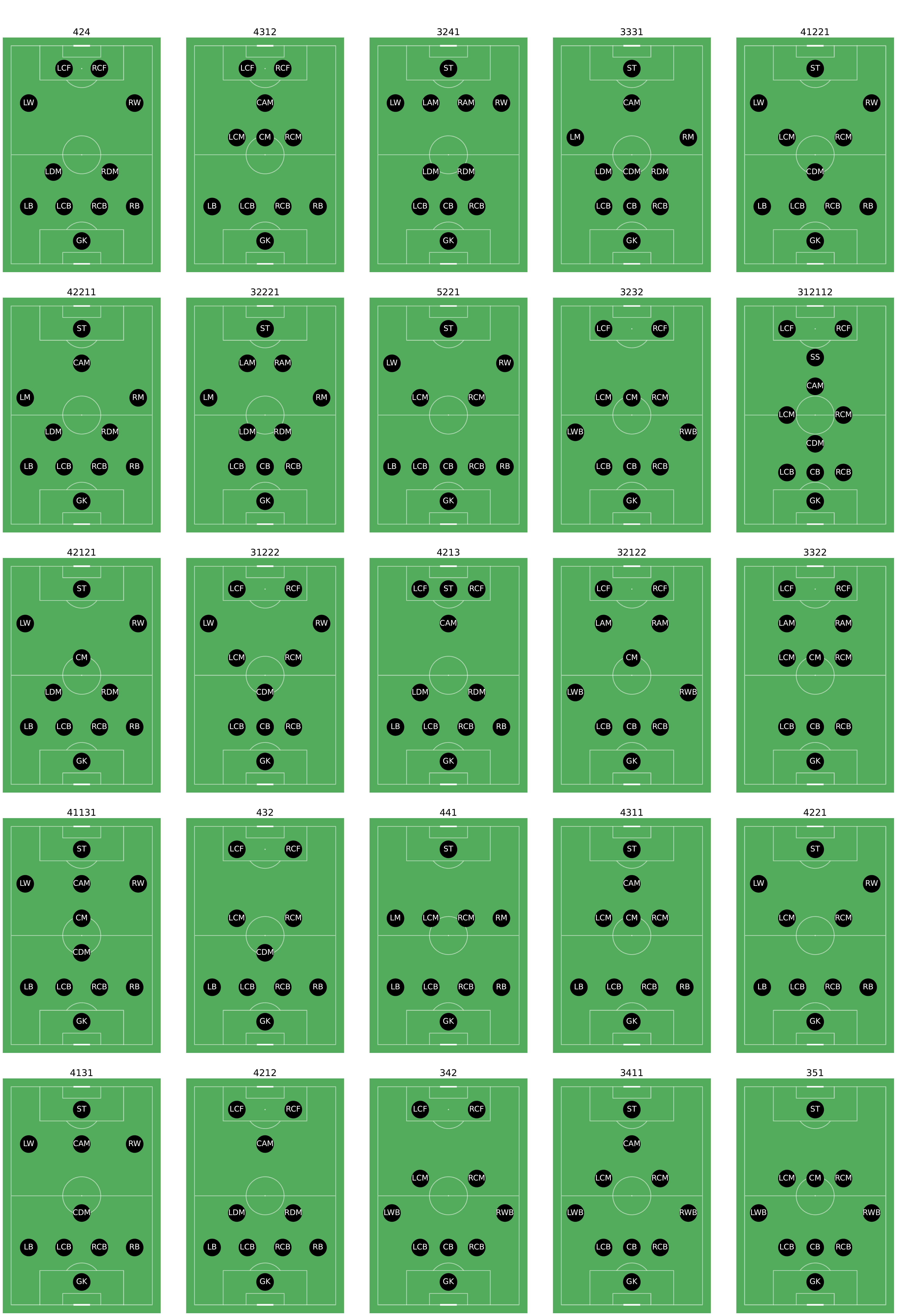}
    \caption{\textit{mplsoccer} Formation Templates 26 - 50}
    \label{fig:positionb}
\end{figure}
\begin{figure}[H]
    \centering
    \includegraphics[width=0.86\linewidth]{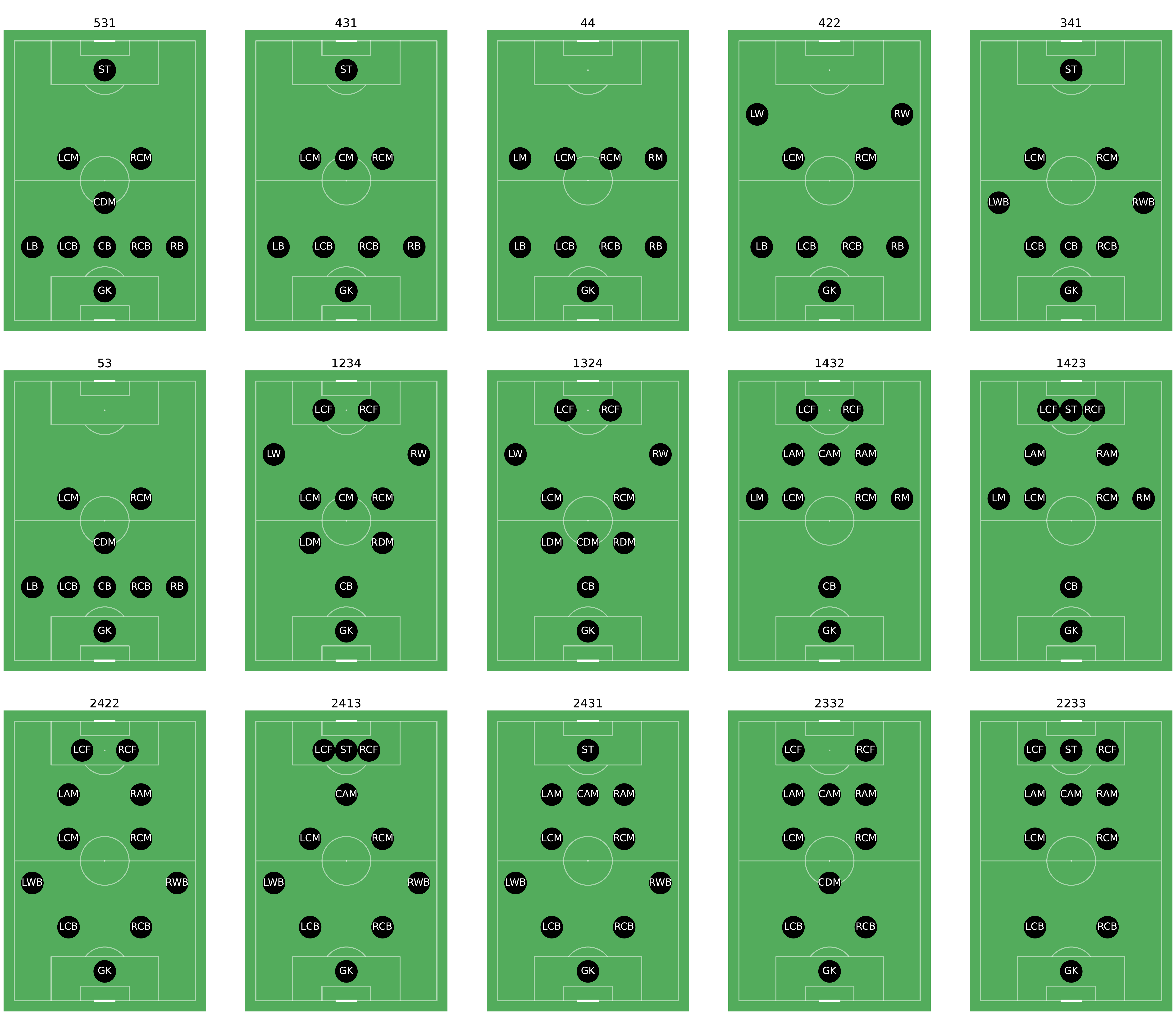}
    \caption{\textit{mplsoccer} Formation Templates 51 - 65}
    \label{fig:positionc}
\end{figure}

\end{appendices}
\newpage
\printbibliography

\end{document}